# Runway Extraction and Improved Mapping from Space Imagery

David A. Noever

PeopleTec, 4901-D Corporate Drive, Huntsville, AL, USA, 35805

**Corresponding author email address**: david.noever@peopletec.com

**Abstract**

Change detection methods applied to monitoring key infrastructure like airport runways represent an important capability for disaster relief and urban planning. The present work identifies two generative adversarial networks (GAN) architectures that translate reversibly between plausible runway maps and satellite imagery. We illustrate the training capability using paired images (satellite-map) from the same point of view and using the Pix2Pix architecture or conditional GANs. In the absence of available pairs, we likewise show that CycleGAN architectures with four network heads (discriminator-generator pairs) can also provide effective style transfer from raw image pixels to outline or feature maps. To emphasize the runway and tarmac boundaries, we experimentally show that the traditional grey-tan map palette is not a required training input but can be augmented by higher contrast mapping palettes (red-black) for sharper runway boundaries. We preview a potentially novel use case (called "sketch2satellite") where a human roughly draws the current runway boundaries and automates the machine output of plausible satellite images. Finally, we identify examples of faulty runway maps where the published satellite and mapped runways disagree but an automated update renders the correct map using GANs.

## 1. Introduction

Airport runways present an attractive challenge for overhead object recognition and modern machine learning. Urban planning, disaster relief, and air safety benefit from tracking runway changes over time (e.g., modeling change detection). In human terms, more than half of all airline accidents occur near airports during takeoff, approach, and landing (Jackman, 2014). Urban planners (such as the International Airports Council) cite airports as the central hub of decaying global infrastructure, with an estimated five-year budget requiring greater than $128 billion investment in the US alone (Baldwin, 2019). A better understanding of current runway condition, either from space or drone imagery, highlights the need for continuous monitoring much like as has been done for highway maps (Bello-Salau, et al., 2014). One motivation for the current work, therefore, is to automate the reversible transformation of unmarked overhead imagery with map-like outlines showing the most current conditions of runways, flight lines, tarmacs, and aprons. In other words, given any runway image, design a method to generate the corresponding map pair and vice versa, as illustrated in Figure 1.

### 1.1 Motivation

A global map identifying the 45,000 small and large places to land an airplane, effectively mirrors the entire earth's landmass with recognizable terrain features (Figure 2, adapted from Davenport, 2013). However, in contrast to the extensive highway and roadway analysis (Chen, et al., 2021), the details of corresponding runway maps have traditionally been low, often just showing a single set of parallel lines to signify the runway. Details like runway damage that might otherwise be identified from updated satellite surveys have not offered a corresponding way to update or evaluate map changes. Kovačič, et al. (2021) noted that sustainable airport management requires real-time evaluation methods, particularly for smaller airports or military bases. Jiang, et al (2020) used deep learning methods to show that aerial (drone) imagery of runway cracks could automate key aspects of inspections. Other cited reasons for not examining the runway integrity problem stem from obfuscation strategies (see examples, Wikipedia, 2021). Notably, the combination of military and civilian runways in entire countries (like Greece) has prompted Google to provide few details in either imagery or maps. Many satellite and map publishers have adopted policies to obfuscate either the image or the map, depending on either the country's or owner's request. Other motivations include algorithmic removal of non-stationary objects like planes from multiple images (effectively emptying the scene of context). In many of these cases, the map is available in full resolution, but the satellite imagery is pixelated beyond recognition (e.g. see Kos International Airport, 36.801622, 27.089944)

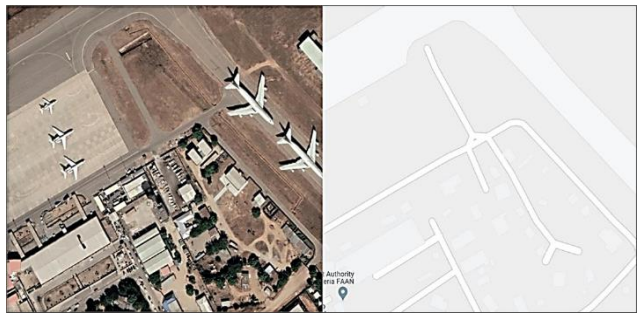

**Figure 1. Example Runway Image and Map Pair.** *Mallam Aminu Kano International Airport, Kano, Nigeria (ICAO Code DNKN).*

### 1.2 Extension of Previous Work

For mapping airport runways algorithmically, the present work examines two popular strategies: Pix2Pix and other generative adversarial networks (GAN) models (Isola, et al., 2017) like CycleGANs. Other image-to-image methods generate synthetic data using CycleGANs (Dou, et al., 2020), Deep Priors (Wang, et al., 2018), Pix2Pix (Isola, et al., 2017), and PatchGANs (Demir, et al., 2018). We address the popular image-to-image (I2I) translation as one concrete example with a practical application (Wang, et al., 2018; Varia, et al., 2018), namely the correction of flawed airport runway maps. We generate and explore a novel dataset for this specialized problem, the translation of an arbitrary airport satellite image into a plausible runway map (Partow, 2017). By specializing in a single feature like airport runways, we ignore the complexities of previous approaches that have combined urban landmarks; ignoring these buildings, roads, and terrain features are tested against the hypothesis that better output might emerge from the more singular focus on one class (Xu, et al. 2018). We hypothesize that a dominant single object type (such as runways) may improve predictive performance. This research systematically controls the map contrast and color saturation to provide new methods for data augmentation and visualization beyond the traditional grey-tan color maps by creating higher contrast black-red maps.

Pix2Pix models have solved several interesting generator-discriminator problems for paired images, including automating the conversion of satellite photos to road maps and vice versa (e.g., sat2map and map2sat transformation, see Isola, et al. 2018). As examples of image-to-image translation and conditional GANs, or cGANs, the output image production depends on the input or paired image. Training urban road data has previously relied on collecting and pairing satellite imagery with low-contrast Carto DB maps (Kang, et al., 2019). These pairs have included complex mixtures of objects, primarily focused on urban scenes with buildings, roads, and parks all combined in the same image. Recent work (Xu, et al. 2018) suggested that research improvements should narrow the object ontology and test for better aesthetic performance.

In addition to narrowing the ontology for runways only, one secondary goal here sought to improve the map contrast examples from the diverse field of GANs include using twin neural networks to spoof satellite images for fake archives and fuse the broad spectral platforms now coming online. The present work applies CycleGANs as one alternative method to generate plausible synthetic runway maps but without relying on precisely paired imagery as conditional requirements. These methods achieve a style transfer between raw images and outline maps as two curated collections without matching location or precise times.

*1.3 Original Contributions*

The research effort offers a novel dataset for training and testing style transfer algorithms like Pix2Pix and CycleGAN. Unlike previous urban settings with mixed object classes, the runway example presents a more uniform case to compare and contrast the algorithmic output. Runways also offer a high-value infrastructure monitoring example where out-of-date or incomplete imagery and maps are not uncommon. We specialize the color palettes of both input and output maps (low-to-high contrast) to examine changes in conditional outputs and enhance visual quality. We identify examples where either the image or map do not match in existing ground truth by human analysis, then compare the generative models for improving on the status quo for incomplete runway mapping. We finally treat the novel use case of converting sketches and rough maps to generate plausible satellite maps. beyond the light grey Carto DB palette and see if a wider color range can improve the Pix2Pix outcome. To explore these issues, we created a novel dataset of 2400 airport locations, mapped in tandem with both Google satellite and map images combined. We systematically enhanced the map color contrast to reduce artifacts and improve appearance for both the "sat2map" and "map2sat" cases.

**2. Methods**

For understanding automated runway maps from images and vice versa, we assembled two datasets. The first one, used for Pix2Pix, requires the map rendering to pair with the exact satellite image in the same location and time. The conditional GAN learns the style transfer features from one to the other, similar to what a colorizing or skeletonizing function might attempt. A Pix2Pix example transforms night and day imagery in pairs. The second dataset, used for CycleGAN, requires no exact pairing of maps and images but instead needs a collection of representative examples for both. The popularity of CycleGAN for doing this kind of unpaired training motivates some classic examples where natural pairs would not normally exist (e.g., horse-to-zebra transformation is not naturally available in the same poses). The primary architectural differences stem from the 2 networks (discriminator-generator) for Pix2Pix but 4 for CycleGANs.

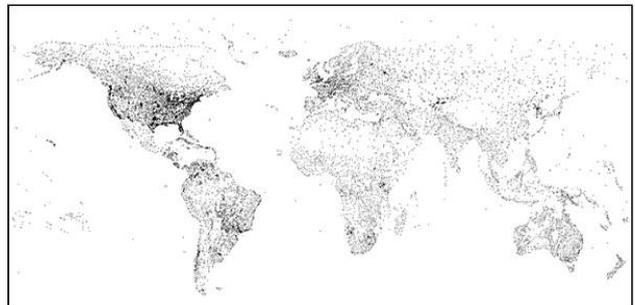

**Figure 2. Global airport distribution (Davenport, 2013).** *Each dot represents an airport and total 45,132 runways ranging from mega-hubs to single dirt roads. The image captures the land borders and global population density using just airport runways (Davenport, 2013).*

*2.1 Dataset Construction and Modification*

We created image pairs of the same location in both map and satellite imagery using the global airport database (9300 latitude-longitude locations, Partow, 2017). The airports ranged across all continents with the highest concentration in the US (552) and Germany (529). We assembled and labeled 2500 images (1200x1200 pixels) from Google Map API both in satellite and map modes of each location centered by latitude and longitude with zoom 18 (approximately 800-1000 feet in altitude). Each square image covers approximately 0.3 square miles. Each image was labeled with the four-letter airport code as location indicators (ICAO, International Civil Aviation Organization).

### 2.2 Dataset Construction and Modification

The runway imagery and maps were scaled to 600x600 pixels, then joined as pairs (1200x600) for Pix2Pix training and validation (Figure 1). Batch image transformations were done using ImageMagick (Still, 2006). For CycleGAN training, the images were reduced further (256x256 pixel) and not joined as pairs but divided into separately labeled folders for A-B and B-A transformations during training (e.g. map-satellite, satellite-map). Zhang, et al. (2020) addressed infrastructure maps using drone aerial imagery and noted the technical challenges of inaccurate or incomplete training data. Our method avoids some of these by relying heavily on the synchrony between satellite and maps already within Google Map API. Where we note obvious differences between the map and image (either because of ground truth changes or poor inputs), we collect those mismatches into inference cases to consider post-training as example use cases. The overlay of place names on maps proved unavoidable based on our API collection method and no attempt was made to remove or obscure them in training or inference datasets.

### 2.3 Training Approach

For both Pix2Pix and CycleGAN methods, the model building benefited from Keras library for deep learning (Chollet, 2018; Brownlee, 2019). The Pix2Pix training for discriminator (D) and generator (G) combinations used the following hyper-parameters: batch size = 1, epochs = 10, base image size = 600x600 pixels in paired A-B sets (600x1200). The CycleGAN training for two sets of D-G combinations used the following hyper-parameters: batch size = 1, epochs = 100, base image size = 256x256 pixels in unpaired A-B sets. The compiled CycleGAN model used the stochastic gradient descent or Adam optimizer (learning rate=0.0002, beta1=0.5, loss_weights=0.5) with mean square error (MSE) loss function. We conducted all training runs using a graphical processing unit (GPU model RTX 5000) with a comparative compute capability=7.0 (Nvidia, 2021). Even when trained on relatively powerful GPU capabilities, the models require several days of training each on dedicated hardware.

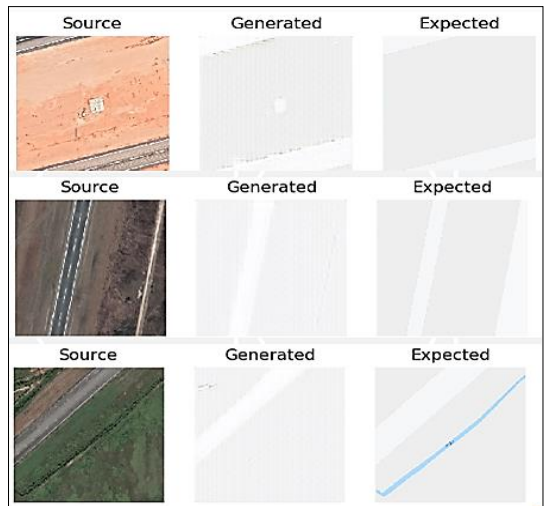

**Figure 3. Pix2Pix model output of plausible maps from satellite images.** *Generated maps are shown compared to ground truth or expected output from Google Maps API.*

## 3. Results

Using the trained GAN models, inference steps ideally generate plausible runway maps from unlabelled satellite images and vice versa. Figure 3 summarizes an example Pix2Pix output of maps generated from satellite imagery.

### 3.1 Map-to-Satellite (Pix2Pix Learning)

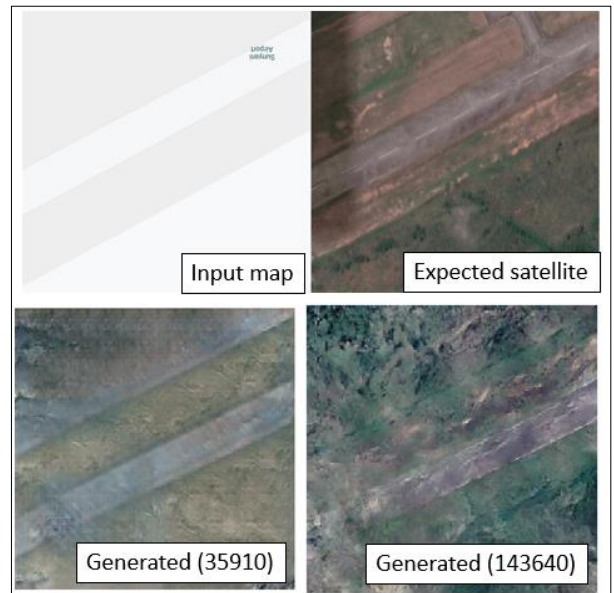

**Figure 4. Conditional GAN (Pix2Pix) model for translating maps into plausible satellite imagery.** *The generated example show learning progress over a number of iterations, with later examples showing more texture and color details.*

For conditional GANs with a required input and an expected outcome (such as Pix2Pix), Figure 4 shows an example output of plausible satellite imagery generated from a simple input runway map. Over many learning steps, the generated satellite imagery includes more texture and color details but does not increase overall realism.

### 3.2 Single Class Comparison

For comparison to traditional multi-class Pix2Pix, Figure 5 shows both a complex urban scene trained to highlight mapped roads and a single-class airport scene trained to show runways. This result supports the hypothesis (Xu, et al. 2018) that reducing the complexity or number of classes in the satellite image might improve map generation. The finer details of roads in Figure 5 appear lost compared to the original satellite image with background buildings and foliage. The simpler runway image generates a reasonable map of coarse but accurate features as needed for navigation or status assessments.

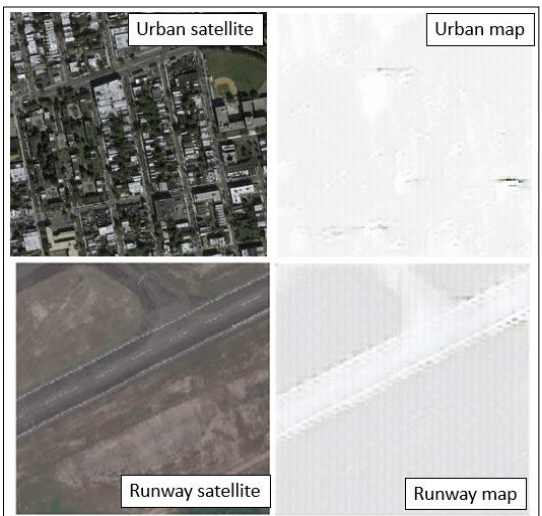

**Figure 5. Comparison of multi- and single-class Pix2Pix generated maps.** *The urban satellite image generates a lower resolution map lacking the features expected and the single-class map shows the simple runway outline as expected.*

### 3.3 CycleGAN Results

Figure 6 shows an example output from both the map and runway inputs. The translated images and map capture a reasonable case, but the maps from CycleGAN generally show less detail and artifacts compared to the maps generated by Pix2Pix paired images. The map-to-satellite transform however performs comparably to Figure 4, Pix2Pix generation.

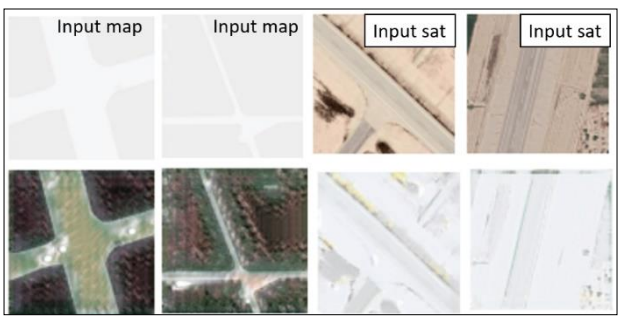

**Figure 8. CycleGAN runway2map and map2runway examples.** *The bottom row is generated from input examples in the top row.*

### 3.4 Comparative GAN Results

For qualitative comparison, the CycleGAN maps appear less convincing in Figure 7 compared to Pix2Pix output. The satellite images for CycleGAN (from map inputs) however show fine details including expected runway color and texture for foliage and open spaces. Although both methods offer reversible translations (map-satellite and satellite-map) the desired runway map features appear superior in Pix2Pix outputs. Depending on the use case, the CycleGAN's ability to generate meaningful color palettes (green) and detailed foliage in "fake" satellite imagery may serve for more lucid descriptions of runway changes.

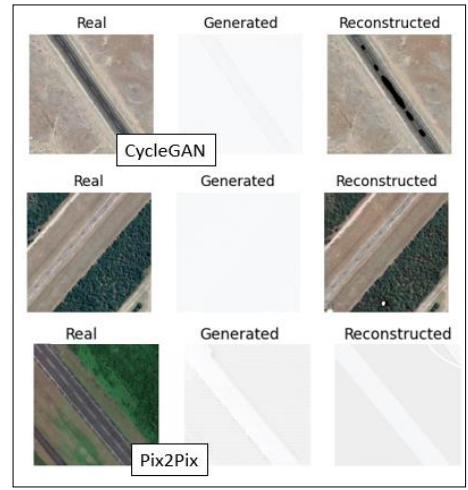

**Figure 6. Comparative output for CycleGAN's four networks vs. Pix2Pix conditional GAN use of 2 networks.** *The satellite-to-map translation for CycleGAN (rows 1-2) loses observable details but compares favorably during image reconstruction. The Pix2Pix model (paired images) provides superior maps.*

### 3.5 High-contrast Map Results

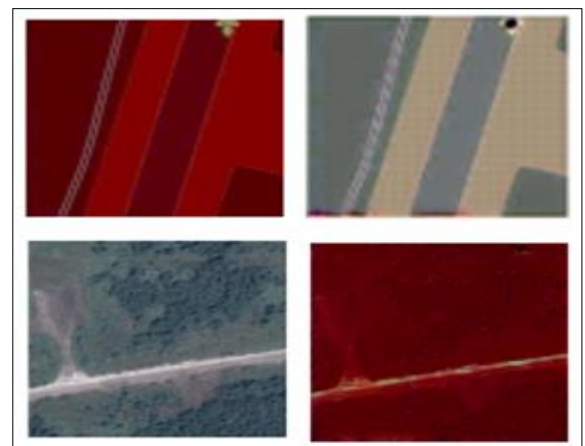

**Figure 7. High-contrast maps (red-black) examples for both A-B and B-A transformation with map2sat (top) and sat2map (bottom).** *Finer foliage details are seen in the high contrast maps for the bottom row compared to the standard tan-gray palette.*

To test whether higher contrast maps can generate better satellite imagery (or vice versa), the input maps were colorized from the dominant tan-grey palette used by Google Maps API to a red-black palette. The CycleGAN model results as trained on high contrast maps are illustrated in Figure 7. While a human geographer might find the high-contrast maps less satisfying, the features of particular interest like runway boundaries, tarmacs, and open space are delineated by high-contrast. The algorithms, both Pix2Pix and CycleGAN, are trainable on any map color palette desired for best results.

### 3.6 Sketch Map Runways

For the interesting use case of a human observer who inspects runway status, then sketches the infrastructure's overall condition, Figure 8 illustrates the image-to-image translation capabilities. The simple maps provide a corresponding plausible satellite image pair. The geometric details of various cross and loop patterns get textured by the Pix2Pix translation. This application represents the first instance of "sketch2satellite" which can be compared to older ground-truth satellite imagery and spawn the corresponding change detection imagery based on human observers, social media posters, and first responders.

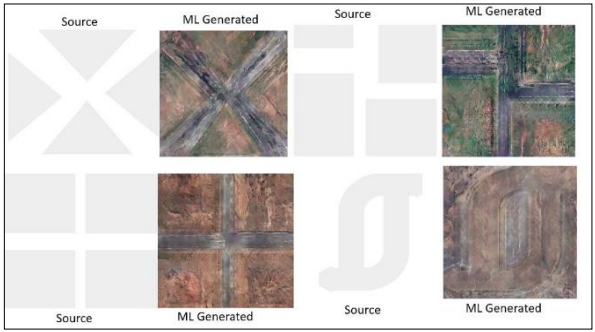

**Figure 9. Example sketched maps and the corresponding generated satellite images.** *The Pix2Pix model renders plausible synthetic satellite data from hand sketches for runways.*

*3.7 Faulty Runway Corrections*

A host of faulty runway maps were identified in the development of the airport dataset. For the sake of clarity, the satellite map generally was assumed to be ground truth, such that running inference models plausibly corrects extra runways or tarmacs misidentified in other public datasets such as Google Maps API. The reverse problem would assume the map is correct and plausibly generate a corresponding satellite image. One implication of this approach is to automate the change detection of outdated maps using recently downloaded orbital images. This change detection scenario effectively automates the most current map always being available for infrastructure monitoring and planning. This similarly has attributes that could be exploited for evaluating runway damage or expected repairs using remote sensing alone.

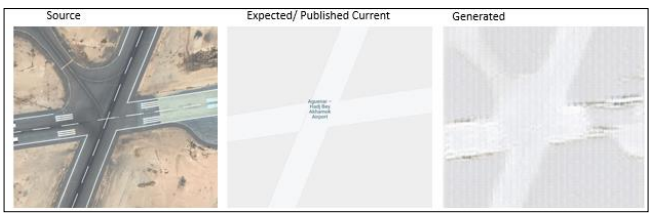

**Figure 11. Faulty Google Map API map for Aguenar Airport, Algeria (DAAT).** *The satellite image shows a five-way intersection, which is correctly captured by the Pix2Pix model but incorrectly shown in the published map.*

*3.8 GAN Artifacts*

A well-known set of artifacts have been noted from previous GAN literature, including smudges, unnatural scenery and phantom landmarks. Figures 3-11 show examples of artifacts that would alert a human expert that the generated image or map is a "fake" or machine generated version compared to the expected ground truth. However, unlike GAN examples with fake faces or counterfeit objects, the generation of runway descriptions (either satellite or maps) adds to the available information and the detectable authenticity proves subordinate to the original intentions. A fake map is assumed to be a simplified overhead image, much like a fake cartoon or painted portrait would be understood as not representing an authentic face.

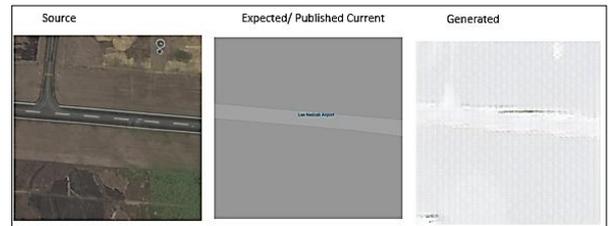

**Figure 10. Faulty Google Map API runway map for Lae Nadzab Airport, Papua New Guinea (AYNZ).** *The satellite image shows a three-way intersection, which is correctly captured by the Pix2Pix model but incorrectly shown in the published map.*

4. Discussion

In addition to curating a novel runway dataset, the present research has 1) demonstrated that single-class image translations can yield lucid maps that highlight the key features like runway boundaries compared to more complex urban environments; 2) identified previously published but incorrect maps using up-to-date satellite imagery to spawn corrected runway boundaries; 3) proposed novel use cases where high-contrast maps or hand-sketches can generate plausible satellite imagery *ab initio* without requiring paired ground truth for predictions.

5. Future Work

An important trend in machine learning has enlarged neural networks, first in-depth with single to multiple layers, then in number or "heads" with single networks to competing ones, and finally in domain expertise with attention and specialized training fields. A key development of multiple networks was the generator-discriminator paradigm that led to generative adversarial networks (GANs). Their remarkable abilities to learn and extend the traditional function approximation or pattern recognition present more creative tasks. Future work builds on and draws inspiration from the rapid growth of small satellites and their concomitant low cost, rapid revisit rates, and public data subscriptions. The research specialization by the spectral quality has broadened considerably with new data fusion opportunities (Linden, et al., 2021). The recent XBD data competition has examined the before-and-after pairings for estimating disaster damage for different grades of building loss (Weber, et al., 2020). Compared to building assessments, the status of re-routing traffic from airports has taken on significance for example in recent relief efforts in Haiti. Further inspiration has followed for the energy industry (pairing EO/IR for estimating thermal efficiency), insurance (damage estimation), and urban planning (satellite to map). Another key development that is just emerging is the multi-domain or multi-modal generalization, where different sensory inputs or perspectives on the same event or object merge into a more interesting and complete picture. There are strong biological analogies for this sensor fusion initiative but the ability to guess a face from a voice or estimate an author from a text snippet represents innovative opportunities for blending multiple domains. Most broadly, if computers achieve expert-level human capabilities for sight, sound and reading, what are the opportunities with creative combinations? A practical implementation step might begin to use the multi-layer, multi-headed networks described here to bridge ever more elaborate creative tasks, sometimes referred to as hallucinating probabilities for sensory translation problems. For Pix2Pix models specifically, the alignment of optical objects like a building with its radar signature (synthetic aperture radar, SAR) offers one intriguing model.

**Acknowledgments**

The author would like to thank the PeopleTec Technical Fellows program for its encouragement and project assistance.

**References**

Baldwin, S. (2019) "Here's why decaying US airports are turning to private money", CNBC, Oct 2, 2019. https://www.cnbc.com/2019/10/02/why-us-airports-are-so-bad.html

Bello-Salau, H., Aibinu, A. M., Onwuka, E. N., Dukiya, J. J., & Onumanyi, A. J. (2014, September). Image processing techniques for automated road defect detection: A survey. In 2014 11th International Conference on Electronics, Computer and Computation (ICECCO) (pp. 1-4). IEEE.

Brownlee, J. (2019). *Generative adversarial networks with python: deep learning generative models for image synthesis and image translation*. Machine Learning Mastery.

Chen, Z., Zhang, Y., Luo, Y., Wang, Z., Zhong, J., & Southon, A. (2021). RoadAtlas: Intelligent Platform for Automated Road Defect Detection and Asset Management. arXiv preprint arXiv:2109.03385.

Chollet, F. (2018). Keras: The python deep learning library. *Astrophysics Source Code Library*, ascl-1806.

Davenport, J. (2013), "The World, Traced by Airport Runways", https://ifweassume.blogspot.com/2013/06/airports-of-world.html including dataset from https://ourairports.com/data/

Demir, U., & Unal, G. (2018). Patch-based image inpainting with generative adversarial networks. *arXiv preprint arXiv:1803.07422*.

Dou, H., Chen, C., Hu, X., Jia, L., & Peng, S. (2020). Asymmetric CycleGAN for image-to-image translations with uneven complexities. Neurocomputing, 415, 114-122.

Isola, P., Zhu, J. Y., Zhou, T., & Efros, A. A. (2017). Image-to-image translation with conditional adversarial networks. In Proceedings of the IEEE conference on computer vision and pattern recognition (pp. 1125-1134).

Jackman, F. (2014) "Nearly Half of Commercial Jet Accidents Occur During Final Approach, Landing", FlightSafety Foundation, https://flightsafety.org/asw-article/nearly-half-of-commercial-jet-accidents-occur-during-final-approach-landing/

Jiang, L., Xie, Y., & Ren, T. (2020). A deep neural networks approach for pixel-level runway pavement crack segmentation using drone-captured images. *arXiv preprint arXiv:2001.03257*.

Kang, Y., Gao, S., & Roth, R. E. (2019). Transferring multiscale map styles using generative adversarial networks. International Journal of Cartography, 5(2-3), 115-141.

Kovačič, B., Doler, D., & Sever, D. (2021). Innovative Business Model for the Management of Airports in Purpose to Identify Runway Damage in Time. Sustainability, 13(2), 613.

Lindén, J., Forsberg, H., Haddad, J., Tagebrand, E., Cedernaes, E., Ek, E. G., & Daneshtalab, M. (2021, October). Curating Datasets for Visual Runway Detection. In *2021 IEEE/AIAA 40th Digital Avionics Systems Conference (DASC)* (pp. 1-9). IEEE.

NVIDIA Specification GPU, Compute Capability, 2021,

Partow, A. (2017) "The Global Airport Database", https://www.partow.net/miscellaneous/airportdatabase/index.html

Senthilnath, J., Varia, N., Dokania, A., Anand, G., & Benediktsson, J. A. (2020). Deep TEC: Deep transfer learning with ensemble classifier for road extraction from UAV imagery. Remote Sensing, 12(2), 245.

Still, M. (2006). *The definitive guide to ImageMagick*. Apress.

Varia, N., Dokania, A., & Senthilnath, J. (2018, November). DeepExt: A Convolution Neural Network for Road Extraction using RGB images captured by UAV. In 2018 IEEE Symposium Series on Computational Intelligence (SSCI) (pp. 1890-1895). IEEE.

Wang, X., Yan, H., Huo, C., Yu, J., & Pant, C. (2018, August). Enhancing Pix2Pix for Remote Sensing Image Classification. In 2018 24th International Conference on Pattern Recognition (ICPR) (pp. 2332-2336). IEEE.

Wang, K., Savva, M., Chang, A. X., & Ritchie, D. (2018). Deep convolutional priors for indoor scene synthesis. *ACM Transactions on Graphics (TOG)*, *37*(4), 1-14.

Weber, E., & Kané, H. (2020). Building disaster damage assessment in satellite imagery with multi-temporal fusion. *arXiv preprint arXiv:2004.05525*.

Wikipedia (accessed 2021), List of satellite map images with missing or unclear data https://en.wikipedia.org/wiki/List_of_satellite_map_images_with_missing_or_unclear_data

Xu, C., & Zhao, B. (2018). Satellite Image Spoofing: Creating Remote Sensing Dataset with Generative Adversarial Networks (Short Paper). In 10th International conference on geographic information science (GIScience 2018). Schloss Dagstuhl-Leibniz-Zentrum fuer Informatik.

Xu, F., Zhang, R., Yang, W., & Xia, G. S. (2019, July). Mental Retrieval of Large-Scale Satellite Images Via Learned Sketch-Image Deep Features. In IGARSS 2019-2019 IEEE International Geoscience and Remote Sensing Symposium (pp. 3356-3359). IEEE.

Zhang, R., Albrecht, C., Zhang, W., Cui, X., Finkler, U., Kung, D., & Lu, S. (2020, August). Map generation from large scale incomplete and inaccurate data labels. In Proceedings of the 26th ACM SIGKDD International Conference on Knowledge Discovery & Data Mining (pp. 2514-2522).

**Author Biographies**

**David Noever** has 31 years of research experience with NASA and the Department of Defense in machine learning and data mining. He received his Ph.D. from Oxford University, as a Rhodes Scholar, in theoretical physics and B.Sc. from Princeton University, *summa cum laude*, and Phi Beta Kappa. His primary research interests center on machine learning, algorithms, data analytics, artificial intelligence, and novel metric generation.